\documentclass[letterpaper]{article} 
\usepackage{aaai20}  
\usepackage{times}  
\usepackage{helvet} 
\usepackage{courier}  
\usepackage[hyphens]{url}  
\usepackage{graphicx} 

\usepackage{mathrsfs}
\usepackage{subfigure}
\usepackage{graphics}
\usepackage{multirow}
\usepackage{multicol}
\usepackage{amsmath}
\usepackage{booktabs}       
\usepackage{amsfonts}       
\usepackage{nicefrac}       
\usepackage{microtype}     
\usepackage{bm}
\usepackage{stfloats}

\usepackage{graphicx}  
\title{Domain segmentation and adjustment for generalized zero-shot learning}
\author{Xinsheng Wang, Shanmin Pang, Jihua Zhu\\  
\textsuperscript {\rm }School of Software Engineering, Xi'an Jiaotong University\\ 
wangxinsheng@stu.xjtu.edu.cn, \{pangsm, zhujh\}@xjtu.edu.cn 
}

\begin{document}

\maketitle

\begin{abstract}
In the generalized zero-shot learning, synthesizing unseen data with generative models has been the most popular method to address the imbalance of training data between seen and unseen classes. However, this method requires that the unseen semantic information is available during the training stage, and training generative models is not trivial. Given that the generator of these models can only be trained with seen classes, we argue that synthesizing unseen data may not be an ideal approach for addressing the domain shift caused by the imbalance of the training data. In this paper, we propose to realize the generalized zero-shot recognition in different domains. Thus, unseen (seen) classes can avoid the effect of the seen (unseen) classes. In practice, we propose a threshold and probabilistic distribution joint method to segment the testing instances into seen, unseen and uncertain domains. Moreover, the uncertain domain is further adjusted to alleviate the domain shift. Extensive experiments on five benchmark datasets show that the proposed method exhibits competitive performance compared with that based on generative models.
\end{abstract}

\section{Introduction}
Zero-shot learning (ZSL) \cite{lampert2009learning} has attracted significant attention in recent years for its ability to recognize newly appeared classes. Compared with traditional recognition methods, ZSL eliminates the restriction whereby the learned model can only be applied to seen classes. Instead, it can face the ever-growing classes in the real world. Despite the revolution of deep convolutional neural networks on traditional image recognition, ZSL remains a challenging problem due to the unavailability of data in the training stage for the scenario during the testing stage.

Similar to the ability of humans to associate different semantic information to infer new instances, ZSL also takes semantic knowledge as auxiliary information to bridge the gap between seen and unseen classes. ZSL aims to learn the knowledge from the set of training classes with corresponding semantic information and labeled data, to predict the testing classes equipped with semantic information. Learning a projection function that facilitates a direct comparison of the visual features and semantic representations in an embedding space, such as the visual feature space, semantic embedding space or a common space, is key for ZSL. Lots of recent works \cite{kodirov2017semantic,liu2018zero} have proved the superiority when the visual feature space is taken as the embedding space on alleviating the hubness problem.

\begin{figure}[t]
\centering
\includegraphics[width=0.9\columnwidth]{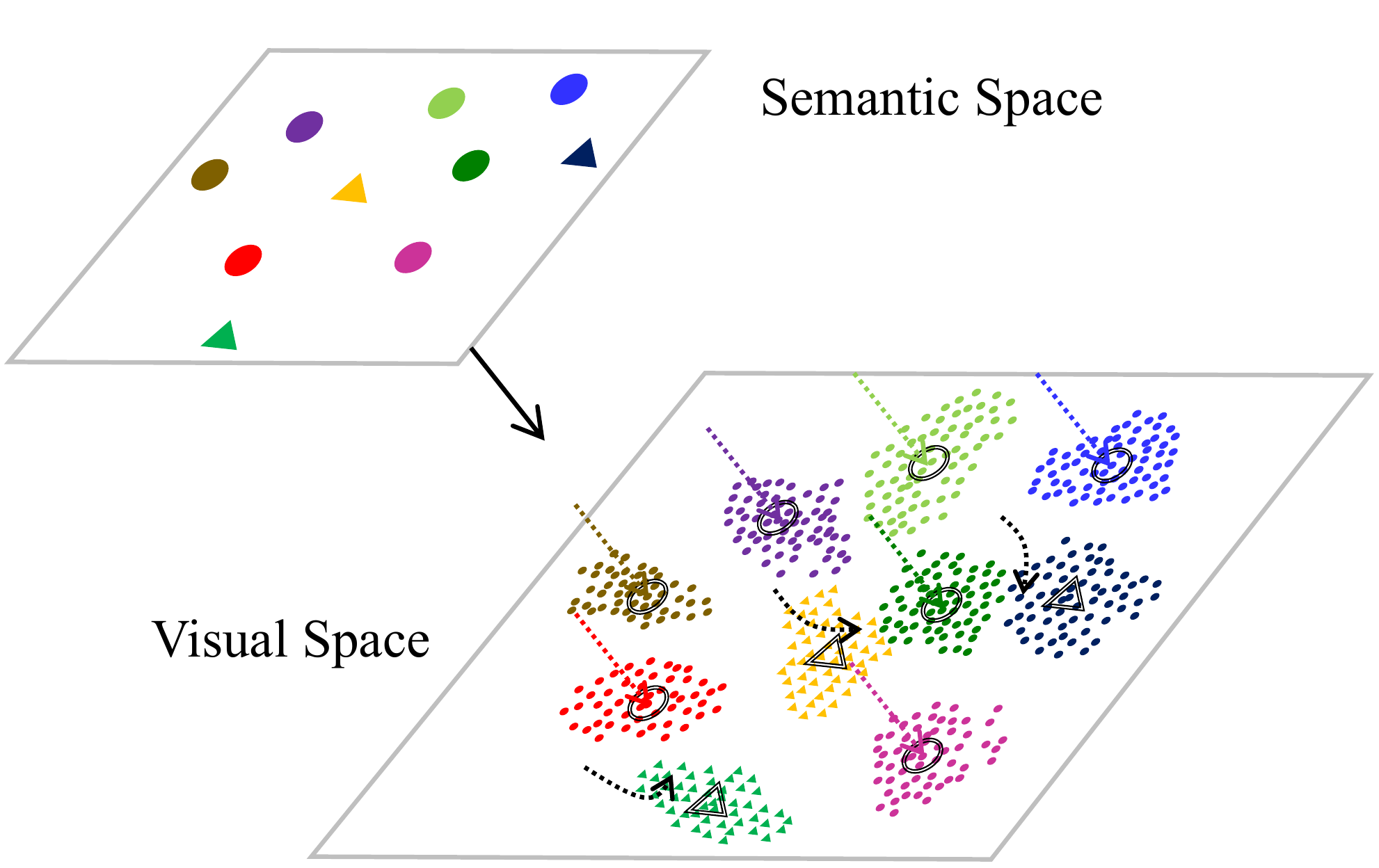} 
\caption{Illustration of the domain shift problem in GZSL. The circles represent the seen classes, and the triangles represent the unseen classes. In the testing stage, unseen semantic embeddings tends to bias seen classes.}
\label{fig1}
\end{figure}

In ZSL, given that the projection for semantic vector embedding is learned from the seen classes, the resulted semantic embedding could be biased towards the seen domains, raising the domain shift problem. The undesired effect of this problem is limited in a conventional ZSL setting, in which only unseen classes are available in the testing stage. However, in the more realistic setting of generalized ZSL (GZSL), the domain shift problem is a serious factor that makes the recognition collapse. As illustrated in Fig. \ref{fig1}, when the visual space is taken as the embedding space, the embedded semantic from the unseen classes tends to bias toward these visual features of the seen classes. To mitigate this issue, the adoption of unseen semantic information to adjust the projection function \cite{jiang2018learning} or the use of the generative model to obtain synthesized unseen features \cite{xian2018feature,schonfeld2019generalized} are two popular strategies that have been adopted in recent GZSL works.

However, on the one hand, allowing the unseen semantic vectors to be available during the training stage breaks the assumption of ZSL that the testing classes are never seen during training. On the other hand, generative models often suffer from instability in the training. In this paper, we introduce a domain segmentation method to address the domain shift problem in GZSL without using any unseen information. In particular, a classifier with softmax output is trained using the seen data. Additionally, the extreme value theory (EVT) is adopted to analyze the distribution of visual features in each seen class. Integrating the softmax responses of the classifier and data distribution resulted by EVT, a method to segment the testing data into seen, unseen and uncertain domains is proposed. Therefore, the seen and unseen domains can avoid the effect from each other during the testing stage. For the uncertain domain, we introduce calibrated stacking to balance the predicted distances from seen and unseen classes.

In summary, our main contributions are as follows:
\begin{quote}
\begin{itemize}
\item An approach that associates the softmax response and EVT results to segment the testing data into seen, unseen and uncertain domains.
\item A method to balance the predicted distances from seen and unseen classes during the testing stage for the uncertain testing samples.
\item New state-of-the-art results for GZSL on five popular benchmarks, including aPY, AwA1, AwA2, CUB, and FLO, without using unseen information. The proposed method also achieves competitive results compared with that using generative models.
\item Comparison and analysis of the effect on GZSL given by synthesized unseen features and calibrated stacking. 
\end{itemize}
\end{quote}


\section{Related work}
In this section, we will first introduce representative related works on ZSL, followed by GZSL and out-of-distribution detection related to domain segmentation.

\subsection{Zero-shot Learning}
In a conventional ZSL task, the training classes and the testing classes are disjointed, and the classification performance is only evaluated on the unseen classes. DAP \cite{lampert2013attribute} is one of the fundamental methods in early ZSL learning researches, which firstly learned the attribute classifier, and then calculated the posterior of a test class for a given instance. Recently, most advances in ZSL are achieved using visual-semantic embedding models that learn a compatibility function between the visual feature space and semantic space \cite{xian2016latent}. This function is then used to map the unseen visual features into the semantic embedding space, or in a contrary direction, followed by the nearest neighbor search for the recognition.

Mapping visual features into the semantic space is an intuitive operation for ZSL, like in \cite{xian2016latent}. However, given that the variance of the data points is likely to be shrunk in this process, this strategy tends to aggravate the hubness problem \cite{shigeto2015ridge}. In contrast, mapping semantic representations into the visual space can reduce the hubness problem. Autoencoder based paradigms make it possible to realize recognition both in the visual space and semantic space. For instance, in SAE \cite{kodirov2017semantic}, a visual feature vector can be projected into the semantic space by the encoder, and the inverse can be realized by the decoder. The following work, LESAE \cite{liu2018zero}, adds the low-rank constraint for the embedding space in the encoder. PSR \cite{annadani2018preserving} considers different relations to preserve the semantic structure in an encoder-decoder process via a multilayer perceptron (MLP) network.

There are also some other recently proposed methods for ZSL problems, such as graph convolutional neural networks-based methods \cite{wang2018zero,kampffmeyer2019rethinking}, production quantization-based method \cite{li2019compressing}, and training a relation net for a comparison between visual features and semantic representations \cite{sung2018learning}. Given that we focus on domain segmentation for GZSL, the simple projection strategy is considered in this paper. Inspired by the outstanding performance of prototype-based recognition algorithms on both conventional classification \cite{yang2018robust} and ZSL tasks \cite{jiang2018learning}, we trained an MLP to connect the semantic representations and learned visual prototypes.

\subsection{Generalized zero-shot learning}
The problem of GZSL is proposed at the very beginning of ZSL works \cite{lampert2009learning}. It is a more reasonable setting for the real-world task, as the testing instances includes both seen and unseen classes. Although almost all existing ZSL approaches can be applied to GZSL, most of them fail to achieve satisfactory results due to the extreme imbalance of the training data between seen and unseen classes. Given that the unseen classes are not considered in the training process, the projected unseen data points are often biased towards seen classes, resulting in the domain shift problem. To address the issue, the work \cite{chao2016empirical} proposed a calibration method, called calibration stacking (CS), which can be used to balance confidence scores of instances from seen and unseen classes. In CDL \cite{jiang2018learning}, semantic vectors from unseen classes were adopted to align the visual-semantic structure. Recently, generating visual features of unseen classes has been the most popular strategy for GZSL. The work \cite{xian2018feature} compared various existing generative models and proposed models to generate unseen visual features via conditional WGAN. Owing to instability in training of GAN-based loss, some works \cite{kumar2018generalized,mishra2018generative} employed conditional variational autoencoders (VAE) for GZSL. The most recent work \cite{xian2019f} combined the advantages of both VAE and GAN with an additional discriminator for GZSL. Instead of generating images or image features, the CADA-VAE \cite{schonfeld2019generalized} generated low-dimensional latent features in a VAE latent space, and the latent distributions from different modalities were aligned in this space. It is noted that these generative methods need access to the semantic information from unseen classes during training. 

Transductive ZSL \cite{verma2017simple,ye2019progressive} is another popular setting strategy to overcome the domain shift problem. Different from the standard ZSL setting, the transductive ZSL implies that the unseen class instances without any labels are accessible during the training stage. With the available unseen class instances, the performance of models on GZSL can be improved effectively for the more useful latent information of unseen classes. 

Although the generative models and transductive learning achieve state-of-the-art performance, to some extent, they breach the restriction that the testing sources should be inaccessible in the training stage. Instead of relying on access to the unseen semantic information or unseen instance features, we tackle the domain shift problem using only the seen resources. As such, the proposed methods are more generally applicable in practice. 

\subsection{Out-of-distribution detection}
Most machine learning systems can only be applied to a close set for which only training classes exist, which is sometimes not practical in real-world deployments. In many applications, reliable detection of the unseen samples is essential to ensure the safety and accuracy of the prediction. Thus, the issue of detecting samples that never appear during the training stage (out-of-distribution detection) has attracted significant attention in recent years.

The work \cite{hendrycks2016baseline} demonstrated that the maximum softmax probability of a pre-trained deep classifier can be used as an unseen example detector. However, a CNN based classifier sometimes will produce incorrect predictions with high predicted class probability even for irrelevant inputs \cite{moosavi2017universal}. The following work \cite{liang2017enhancing} improved the performance by pre-processing input data with adversarial perturbations to make the maximum softmax probability more discriminative between anomalies and in-distribution examples. In \cite{lee2017training}, synthesized samples acting as out-distribution were used in the training process. Instead of using synthesized samples, OE \cite{hendrycks2018deep} leveraged auxiliary dataset of outliers to improve deep anomaly detection. Moreover, extreme value theory (EVT), which is a probabilistic distribution method instead of a single predicted score, exhibits outstanding performance in related fields. In OpenMax \cite{bendale2016towards}, an extreme value model was used to update the penultimate vector for open-set recognition. Most recently, \cite{oza2019deep} used the EVT to model the tail of the reconstruction error distribution from the known class for open-set recognition. 

In ZSL, the unseen classes in the testing stage are outliers, in contrast with the seen classes. Therefore, we propose to segment the seen and unseen classes with out-of-distribution detection methods. However, for GZSL, we are not concerned with detecting unseen examples, but achieving powerful performance on GZSL task. Thus, besides segmenting the testing examples into seen and unseen, we introduce the third domain, namely, the uncertain domain, to reduce error during the segmentation process. Instead of making use of generative models or auxiliary datasets for the detection of unseen classes, only the training data of the seen classes are adopted. 

\section{Approach}
\subsection{Preliminaries}
The training set is given as $\left\{ {x_i^{\left( s \right)},y_i^{\left( s \right)},l_i^{\left( s \right)}} \right\}$, where $x_i^{\left( s \right)} \in {{\cal V}_s}$ is a $d$-dimensional feature of an image in the visual space, $y_i^{\left( s \right)} \in {{\cal Y}_s}$ is the semantic vector of the seen class $i$, and the ${l_i} \in {L_s}$ is the corresponding label. Similarly, the testing set also includes a set of visual features ${{\cal V}_u}$, semantic vectors ${{\cal Y}_u}$ and the corresponding label ${L_u}$. In the conventional ZSL, the goal is to learn a classifier to predict the label of ${x_i} \in {{\cal V}_u}$ during the testing stage. For GZSL, not only instances from unseen classes but also from seen classes are included, and the semantic representations and corresponding labels are also given by ${{\cal Y}_s} \cap {{\cal Y}_u}$ and ${L_s} \cap {L_u}$, respectively. In an embedding-based GZSL learning approach, the goal is to learn a mapping function, such as a semantic-to-visual function $\psi :{\cal Y} \to {\cal V}$, and the label of a testing image $x$ can be predicted by
\begin{equation}
    l^* = \mathop {\arg \min }\limits_{l \in L} {\left\| {x - \psi \left( {{y_l}} \right)} \right\|^2}
\end{equation}

\subsection{Overview of the approach}
\begin{figure*}[t]
\centering
\includegraphics[width=0.9\textwidth]{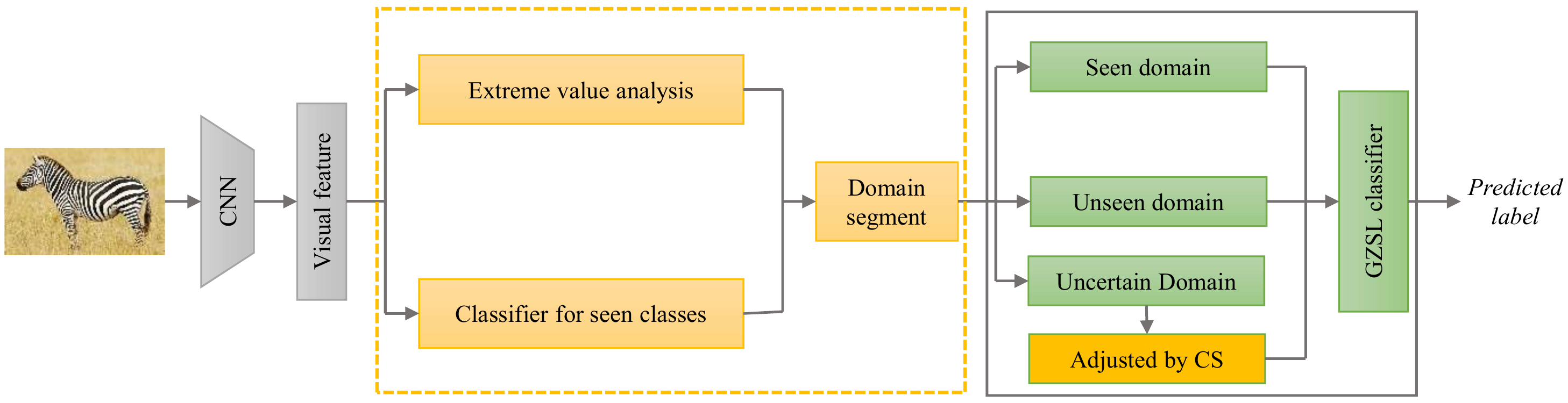} 
\caption{Architecture of the proposed approach. Firstly, we unify the EVT and softmax confidence score to predict the domain of a testing instance (orange box). Then, the prediction of the testing instance is applied within its domain by the trained embedding function for GZSL (green box). For the uncertain domain, CS is applied for the further domain adjustment.}
\label{fig2}
\end{figure*}
To address the domain shift problem in GZSL, the key idea of this paper is to segment the testing instances into different domains. Specifically, we first train a classifier for the seen classes. In addition, EVT is adopted. Then, the output of softmax and the probability distribution produced by EVT are unified to segment the testing instances into three domains, including the seen domain, unseen domain, and uncertain domain. Finally, the instances in the different domains are further predicted by GZSL classifier. Additionally, we take the embedding-based method as GZSL classifier. The architecture of the proposed method is illustrated by Fig. \ref{fig2}.

\subsection{Classifier for seen classes}
Let the neural network $  \bm{f}  = \left( {{f_1},...,{f_c},...,{f_p}} \right)$ be the model to classify $p$ seen classes. Given a visual feature ${x_i}$, the probability distribution over the seen classes produced by the softmax function is
\begin{equation}
    {p_c}\left( {{x_i}} \right) = \frac{{\exp \left( {{f_c}\left( {{x_i}} \right)} \right)}}{{\sum\limits_{m = 1}^p {\exp \left( {{f_m}\left( {{x_i}} \right)} \right)} }}
    \label{eq:OD_softmax}
\end{equation}
However, the deep classifier trained by the original softmax probability tends to have an overfitted confidence for the seen classes, and high scores are obtained even for the unseen classes. To mitigate the overfitting problem and to facilitate better discrimination for unseen class detection, we apply temperature calibration (TC) \cite{hinton2015distilling} to Eq. \ref{eq:OD_softmax} as follows:
\begin{equation}
    p_{c}^{\prime}\left( {{x_i}} \right) = \frac{{\exp \left( {{{{f_c}\left( {{x_i}} \right)} \mathord{\left/
 {\vphantom {{{f_c}\left( {{x_i}} \right)} \tau }} \right.
 \kern-\nulldelimiterspace} \tau }} \right)}}{{\sum\limits_{m = 1}^p {\exp \left( {{{{f_m}\left( {{x_i}} \right)} \mathord{\left/
 {\vphantom {{{f_m}\left( {{x_i}} \right)} \tau }} \right.
 \kern-\nulldelimiterspace} \tau }} \right)} }}
 \label{eq:OD_softmax_CS}
\end{equation}
where $\tau$ is the temperature scale. It should be noted that the TC is only adopted for the training stage. During the testing stage, the predicted confidence score is given by Eq. \ref{eq:OD_softmax}. Cross-entropy loss is employed to train the classier. With the learned classifier, the confidence score for a given example $x_i$ is calculated as follows:
\begin{equation}
    {h_i} = \mathop {\max }\limits_c \frac{{\exp \left( {{f_c}\left( {{x_i}} \right)} \right)}}{{\sum\limits_{m = 1}^p {\exp \left( {{f_c}\left( {{x_i}} \right)} \right)} }}
\end{equation}

\subsection{Extreme value distribution}
Although the TC is applied in the classifier training, some unseen examples still obtain high confidence scores with the softmax output. Moreover, the accuracy of the trained classifier is limited because the CNN used to extract visual features is not fine-tuned. Thus, instead of using confidence scores of the softmax output, the extreme distribution is analyzed using the distances between instances and class centroids in the visual space. For a class $c$, if there are $N$ instances in the training set, the visual centroid of this class is calculated as follows:
\begin{equation}
    {z_c} = \frac{1}{N}\sum\limits_{i = 1}^N {\left( {\left. {{x_i}} \right|{x_i} \in Class\;c} \right)} 
    \label{eq:class_center}
\end{equation}
The distances between the visual centroid and all instances belonging to the same class are calculated, and then the tails of $n$ largest distances are chosen to get the probability distribution that follows the Cumulative Distribution Function (CDF) of Weibull distribution as follows:
\begin{equation}
    {P_c} = 1 - \exp \left( { - {{\left( {\frac{{{d_{i,c}} - {v_c}}}{{{\lambda _c}}}} \right)}^{{k_c}}}} \right)
\end{equation}
where ${d_{i,c}}$ is the distance between sample $x_i$ and the visual centroid of class $c$. The parameters $v_c$, $\lambda _c$ and $k_c$ are the location, scale, and shape of the Weibull distribution respectively, and they are estimated using the library provided by \cite{scheirer2012multi}. Given a sample $x_i$, if the $P_c$ is large and close to 1, it is out of class $c$. In contrast, if the value is small, it belongs to class $c$. The estimation can be expressed as follows:     
\begin{equation}
    {O_{c}} = \left\{ {1\left| {{P_c} > {\alpha _o}} \right.} \right\}
\end{equation}
\begin{equation}
    {I_{c}} = \left\{ {\left. 1 \right|{P_c} < {\alpha _{in}}} \right\}
\end{equation}
where ${\alpha _o}$ and ${\alpha _{in}}$ are thresholds for out of class $c$ and in class $c$, respectively.

\subsection{Domain segmentation}
Given a sample $x_i$, classes are sorted in descending order according to the confidence scores obtained by Eq. \ref{eq:OD_softmax} as $c_1$, $c_2$,...,$c_p$. The domains can be segmented as
\begin{equation}
\text {domain }=\left\{\begin{array}{cc}{\text { Seen }} & {h_{c_1}>\beta_{\text {in}} \& I_{c_1}=1} \\ {\text {Unseen}} & {h_{c_1}<\beta_{\text {out}} \& \sum_{i=1}^{k} O_{c_i}=k} \\ {\text {Uncertain}} & {\text {others}}\end{array}\right.
\label{eq:DS}
\end{equation}
where ${\beta _{in}}$ and ${\beta _{out}}$ are thresholds for in the distribution and the out of the distribution, respectively. $h_{c_i}$ means the top-i softmax score.
\subsection{Embedding}
For simplicity, we train an MLP mapping model with semantic representations and visual prototypes. Given the discrete distribution of instances in visual space, a prototype should be learned for each class of visual features. Following \cite{wang2019visual}, we initialize visual prototypes with the visual centroid of each class expressed by Eq. \ref{eq:class_center}. Then, the visual prototypes are trained with the following loss function:
\begin{equation}
\mathcal{L}_{p}=-\sum_{i=1}^{N_{s}} \sum_{j=1}^{p} \mathbb{I}_{i, j} \log \left(\hat{m}_{i, j}\right)
\end{equation}
where $\mathbb{I}_{i, j}$ is an indicator function for label $x_i$, and $N_s$ is the number of training instances. ${\hat m_{i,j}}$ is given by
\begin{equation}
    {\hat m_{i,j}} = \frac{{\exp \left( {{m_{i,j}}} \right)}}{{\sum\limits_{k = 1}^p {\exp } \left( {{m_{i,k}}} \right)}}
\end{equation}
where ${m_{i,j}}$ is the similarity of the visual feature $x_i$ with the visual prototype of class $j$. In this case, we take the inner product between the visual features and the visual prototypes as the similarities. With the learned prototypes, the object function for the embedding can be 
\begin{equation}
    \mathop {\min }\limits_{_\psi } \sum\limits_{i = 1}^p {{{\left\| {\psi \left( {y_i^{(s)}} \right) - \hat {z}_i} \right\|}^2}} 
\end{equation}
where $\psi (.)$ is an MLP network working to embed the semantic vectors into the visual space. $\hat {z}_i$ is the learned prototype of class $i$. The loss function for training the embedding function is given by
\begin{equation}
\begin{array}{l}
\begin{aligned}
\mathcal{L}_{e}=\sum_{i=1}^{p}\left\|\left(W_{2} f\left(W_{1} y_{i}^{(s)}\right)-\hat {z}_i \right)\right\|^{2}\\
+\lambda_{e}\left(\left\|W_{1}\right\|^{2}+\left\|W_{2}\right\|^{2}\right)
\end{aligned}
\end{array}
\end{equation}
where $W_1$ and $W_2$ are parameters of the first and second layer of the MLP network, respectively. $f\left(  \cdot  \right)$ is the ReLU algorithm. ${\lambda _e}$ is the hyperparameter that acts as a weight for the regularization loss of the parameters. With the learned mapping function $\psi (.)$, the recognition of a testing sample $x_i$ is given by
\begin{equation}
l^*=\left\{\begin{array}{l}{\underset{l \in L_{u}}{\operatorname{argmin}}\left\|x_{i}-\psi\left(y_{l}\right)\right\|^{2} \quad \text { Unseen domain }} \\ {\underset{l \in L_{s}}{\operatorname{argmin}}\left\|x_{i}-\psi\left(y_{l}\right)\right\|^{2} \quad \text { Seen domain }} \\ {\underset{l \in {L_u} \cap {L_s}}{\operatorname{argmin}}\left\|x_{i}-\psi\left(y_{l}\right)\right\|^{2} \cdot J \quad \text { Uncertain domain }}\end{array}\right.
\end{equation}
where $J$ is used for the calibration stacking:
\begin{equation}
J=\left\{\begin{array}{ll}{\gamma} & {l \in L_{s}} \\ {1} & {l \in L_{u}}\end{array}\right.
\end{equation}
and $\gamma$ is a parameter that is larger than 1 to balance the distance from the seen and unseen classes.

\section{Experiments}
In this section, the experimental evaluation of the proposed method is presented, and the model is analyzed using ablation study. In addition, the effectiveness of synthesized features of the generative models in GZSL is investigated.

\subsection{Dataset and setting}
Five commonly used benchmark datasets, including aPY \cite{farhadi2009describing}, AwA1 \cite{lampert2009learning}, AwA2 \cite{xian2016latent}, CUB \cite{wah2011caltech}, and FLO \cite{nilsback2008automated} were adopted. Among them, aPY is a small-scale coarse-grained dataset with a total of 32 classes and 15339 images. AwA1 and AwA2 are coarse-grained datasets, with 50 classes of animals. These datasets are medium-scale in terms of the total number of images, i.e., 30475 and 37322, respectively. CUB is a medium-scale dataset containing 200 classes and 11788 images. FLO is a fine-grained dataset with 8189 images from 102 different classes of flowers. For a fair comparison, we used the zero-shot splits proposed by \cite{xian2018zero} for aPY, AwA1, AwA2, and CUB. The split of FLO follows the standard split given by \cite{reed2016learning}.

For the visual features, we also followed the setting in the work \cite{xian2018zero}. Specifically, the visual features are 2048-dim vectors that were extracted from the entire image using the 101-layered ResNet, and ResNet was pre-trained on ImageNet 1K without fine-tuning. As semantic representations, attributes were adopted for datasets aPY, AwA1, AwA2, and CUB. For FLO, we selected 1024-dim features extracted from CNN-RNN as in \cite{xian2018feature}. The attribute vector dimensions of aPY, AwA1, AwA2, and CUB are 64, 85, 85, and 312, respectively. 

Average per-class top-1 accuracy was adopted for the evaluation of single-label image classification accuracy. In the setting of GZSL, both seen and unseen classes appear during the testing stage. To obtain the overall evaluation of performance for both seen and unseen classes, we followed \cite{xian2018zero} to use the harmonic mean of the seen and unseen accuracy. Let $ACC_{tr}$ and $ACC_{ts}$ denote the recognition accuracy of images from seen and unseen classes respectively, the harmonic mean H of the seen and unseen accuracy is defined as
\begin{equation}
 H = \frac{{2AC{C_{tr}} \times AC{C_{ts}}}}{{AC{C_{tr}} + AC{C_{ts}}}} 
\end{equation}

\subsection{Implementation details}
The classifier in the domain segmentation and that in GZSL were both implemented with PyTorch. In the EVT process, tails with a size of 20 with the largest distances from the corresponding visual centroids were adopted. The value of $k$ in Eq. \ref{eq:DS} was 3, which means that an example would not be segmented into the unseen domain as long as it belongs to any class of the top-3 most similar classes. In the training of the embedding function, the learned visual prototypes and parameters for the embedding function were updated alternately. The details of the training parameters and other parameters can be seen in our public source code.

\subsection{Comparison with the state-of-the-art}

\begin{table*}[hbp]
\centering
\caption{Generalized zero-shot results on aPY, AwA1, AwA2, CUB, and FLO. The results are measured in average per-class top-1 accuracy (\%). $\dagger$ denotes using generative models conditioned on unseen semantic information.}
\setlength{\tabcolsep}{0.55mm}
\label{tab:my-table}
\begin{tabular}{l|lll|lll|lll|lll|lll}
\toprule
 & \multicolumn{3}{c|}{aPY} & \multicolumn{3}{c|}{AwA1} & \multicolumn{3}{c|}{AwA2} & \multicolumn{3}{c|}{CUB} & \multicolumn{3}{c}{FLO} \\ \hline
\multicolumn{1}{c|}{Method} & ts     & tr     & H      & ts      & tr     & H      & ts      & tr     & H      & ts     & tr     & H      & ts     & tr     & H     \\ \hline
SSE \cite{zhang2016zero}                & 0.2    & 78.5   & 0.4    & 7.0     & 80.5   & 12.9   & 8.1     & 82.5   & 14.8   & 8.5    & 46.9   & 14.4   & ---    & ---    & ---   \\
LATEM \cite{xian2016latent}              & 0.1    & 73.0   & 0.2    & 7.3     & 71.7   & 13.3   & 11.5    & 77.3   & 20.0   & 15.2   & 57.3   & 24.0   & ---    & ---    & ---   \\
SAE \cite{kodirov2017semantic}               & 0.4    & $\mathbf{80.9} $  & 0.9    & 1.8     & 77.1   & 3.5    & 1.1     & 82.2   & 2.2    & 7.8    & 54.0   & 13.6   & ---    & ---    & ---   \\
RN \cite{sung2018learning}                & ---    & ---    & ---    & 31.4    & $\mathbf{91.3}$   & 46.7   & 30.0    & $\mathbf{93.4}$   & 45.3   & 19.6   & 54.0   & 13.6   & ---    & ---    & ---   \\
LESAE \cite{liu2018zero}  & 12.7   & 56.1   & 20.1   & 19.1    & 70.2   & 30.0   & 21.8    & 70.6   & 33.3   & 19.1   & 70.2   & 30.0   & ---    & ---    & ---   \\
PSR \cite{annadani2018preserving}                & 13.5   & 51.4   & 21.4   & ---     & ---    & ---    & 20.7    & 73.8   & 32.3   & 24.6   & 54.3   & 33.9   & ---    & ---    & ---   \\
PREN \cite{ye2019progressive}                & ---    & ---    & ---    & ---     & ---    & ---    & 32.4    & 88.6   & 47.4   & 35.2   & 55.8   & 43.1   & ---    & ---    & ---   \\
MLSE \cite{ding2019marginalized}                & 12.7   & 74.3   & 21.7   & ---     & ---    & ---    & 23.8    & 83.2   & 37.0   & 22.3   & 71.6   & 34.0   & ---    & ---    & ---   \\
VSE \cite{zhu2019generalized}                 & 24.5   & 72.0   & 36.6   & ---     & ---    & ---    & 41.6    & 91.3   & 57.2   & 33.4   & $\mathbf{87.5}$   & 48.4   & ---    & ---    & ---   \\
PQZSL \cite{li2019compressing}               & 27.9   & 64.1   & 38.8   & ---     & ---    & ---    & 31.7    & 70.9   & 43.8   & 43.2   & 51.4   & 46.9   & ---    & ---    & ---   \\ \hline
Our                         & $\mathbf{36.0}$  & 56.8   & $\mathbf{44.1}$   & $\mathbf{61.2}$    & 71.7   & $\mathbf{66.0}$   & $\mathbf{58.7}$    & 76.7   & $\mathbf{66.5}$   & $\mathbf{47.7}$   & 53.3   & $\mathbf{50.3}$   & 58.2   & 72.2   & 64.5  \\ \hline
f-CLSWGAN \cite{xian2018feature} $\dagger$          & ---    & ---    & ---    & 57.9    & 61.4   & 59.6   & ---     & ---    & ---    & 43.7   & 57.7   & 49.7   & 59.0   & 73.8   & 65.6  \\
LisGAN \cite{li2019leveraging} $\dagger$            & ---    & ---    & ---    & 52.6    & 76.3   & 62.3   & ---     & ---    & ---    & 46.5   & 57.9   & 51.6   & 57.7   & 83.8   & 68.3  \\
GDAN \cite{huang2019generative} $\dagger$             & 30.4   & 75.0   & 43.4   & ---     & ---    & ---    & 32.1    & 67.5   & 43.5   & 39.3   & 66.7   & 49.5   & ---    & ---    & ---   \\
CADA-VAE \cite{schonfeld2019generalized} $\dagger$           & ---    & ---    & ---    & 57.3    & 72.8   & 64.1   & 55.8    & 75.0   & 63.9   & 51.6   & 53.5   & 52.4   & ---    & ---    & ---   \\
f-VAEGAN-D2 \cite{xian2019f} $\dagger$       & ---    & ---    & ---    & ---     & ---    & ---    & 57.6    & 70.6   & 63.5   & 48.4   & 60.1   & 53.6   & 56.8   & 74.9   & 64.6  \\ \bottomrule
\end{tabular}
\label{tab:GZSL}
\end{table*}

Apart from state-of-the-art methods that do not exploit any generative models, some recently proposed methods with augmented unseen features are also compared in this section. Table \ref{tab:GZSL} presents the result of GZSL for five datasets. In the table, ts and tr refer to $ACC_{ts}$ and $ACC_{tr}$, respectively. The target labels for the evaluation of both ts and tr are full labels, including seen and unseen labels. The comprehensive evaluation of GZSL is given by the harmonic mean (H) of tr and ts.

Compared with previous methods that do not exploit any generative model, the proposed method achieves the new state-of-the-art performance. As shown in Table \ref{tab:GZSL}, our method outperforms these methods for all datasets. Specifically, the proposed method is superior to the most competitive method VSE \cite{zhu2019generalized}, which had the best performance on the datasets AwA2 and CUB, by 9.3\% and 1.9\%, respectively. For dataset aPY, the previous best result achieved by PQZSL \cite{li2019compressing} is 38.8\%, compared with which, the gain brought by the proposed method is 5.3\%. Our method also achieves the best performance for the dataset AwA1, and the improvements are 19.3\% and 30\% compared with the recently proposed methods RN \cite{sung2018learning} and LESAE \cite{liu2018zero}.

Additionally, the pleasant surprise is that the proposed method even exhibits competitive performance compared with those methods based on generated samples. Specifically, the proposed method is superior to all these methods when a comparison is made using datasets AwA1 and AwA2. The gains are 1.9\% and 2.6\% for these two datasets, respectively, compared with CADA-VAE \cite{schonfeld2019generalized}, which yielded the best results. In addition, the proposed method also achieves better performance than GDAN \cite{huang2019generative} by 0.7\% on the dataset aPY. For the dataset CUB, the proposed method also outperforms f-CLSWGAN \cite{xian2018feature} and GDAN \cite{huang2019generative}. Although the current method does not outperform these based on synthesized features for dataset FLO, it exhibits similar performance with f-CLSWGAN \cite{xian2018feature} and f-VAEGAN-D2 \cite{xian2019f}. It is worth noting that all these methods based on generative models were conditioned on the unseen semantic information, which implies that they must access the semantic representations from unseen classes. In contrast, the proposed method neither needs the unseen semantic information nor takes any augmented samples.

\subsection{Model analysis}
In this section, AwA1 and FLO were taken as examples for model analysis. The effectiveness of domain segmentation (DS) and calibrated stacking (CS) is presented in Table \ref{tab:ablation}. The baseline is the embedding function without DS and CS. Baseline + CS indicates that CS was implemented on all testing samples. As shown in this table, the proposed DS has a competitive performance compared with CS. Specifically, on AwA1, the gains of H due to CS and DS are 8.6\% and 7.2\% respectively. The implementations of CS and DS are better than the baseline by 11.3\% and 12.9\%, respectively for FLO. Although CS exhibits outstanding performance on GZSL, the implementation of CS on all the test samples results in excessive adjustment for some unseen classes. This issue can be alleviated if the instances of high confidences for seen (seen domain) or unseen (unseen domain) classes are free from CS. As presented in Table \ref{tab:ablation}, the combination of CS and DS achieves higher accuracy and the gains of H on datasets AwA1 and FLO are 9.3\% and 14.3\%, respectively.

\begin{table}[h]
\centering
\caption{Ablation performance of the proposed method on GZSL. The results are measured in average per-class top-1 accuracy (\%).
}
\label{tab:my-table2}
\setlength{\tabcolsep}{1.3mm}
\begin{tabular}{l|lll|lll}
\toprule
                   & \multicolumn{3}{c|}{AwA1} & \multicolumn{3}{c}{FLO} \\ \hline
Method             & ts      & tr     & H      & ts     & tr     & H     \\ \hline
Baseline           & 43.7    & 79.3   & 56.3   & 37.0   & 78.4   & 50.2  \\
Baseline + CS      & 57.9    & 73.9   & 64.9   & 51.6   & 76.0   & 61.5  \\
Baseline + DS      & 54.2    & 76.6   & 63.5   & 53.9   & 76.2   & 63.1  \\
Baseline + DS + CS & 61.2    & 71.7   & 66.0   & 58.2   & 72.4   & 64.5  \\ \bottomrule
\end{tabular}
\label{tab:ablation}
\end{table}

\begin{figure}[h]
\centering
\subfigure{
\begin{minipage}{0.54\columnwidth}
   \includegraphics[width=0.98\columnwidth]{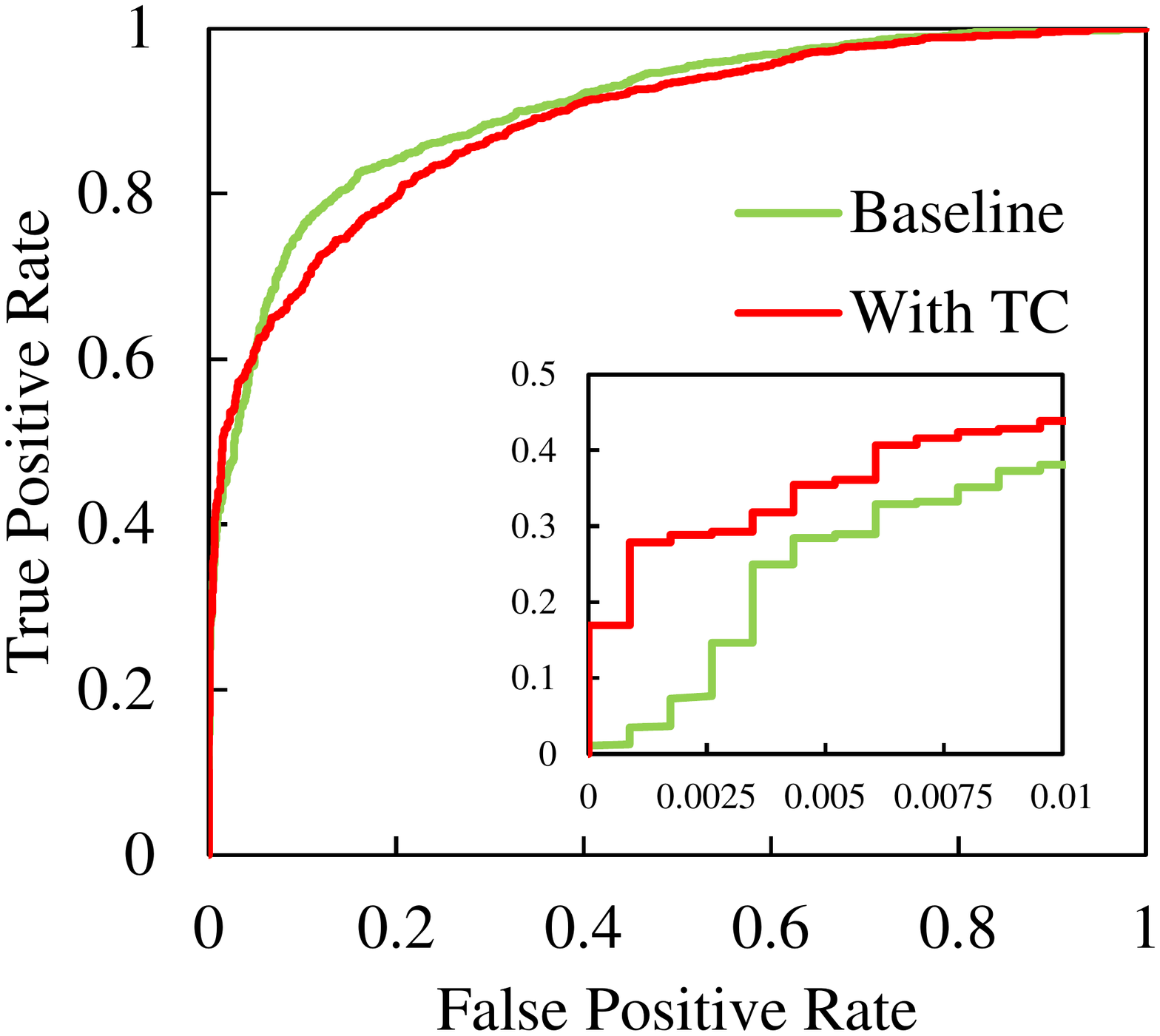}
  \end{minipage}}
\subfigure{ 
\begin{minipage}{0.42\columnwidth}
   \includegraphics[width=0.9\columnwidth]{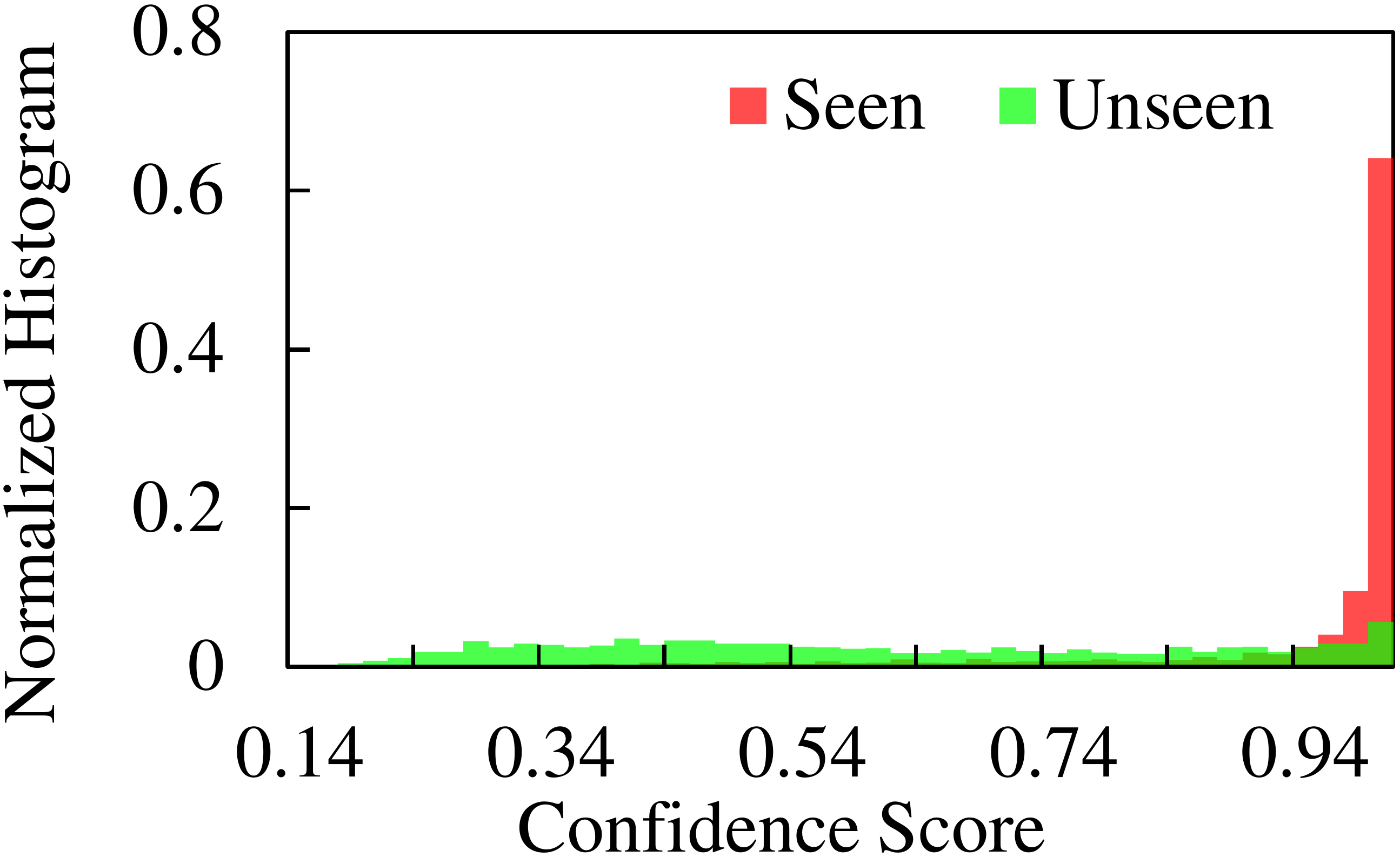}\\
   \includegraphics[width=0.9\columnwidth]{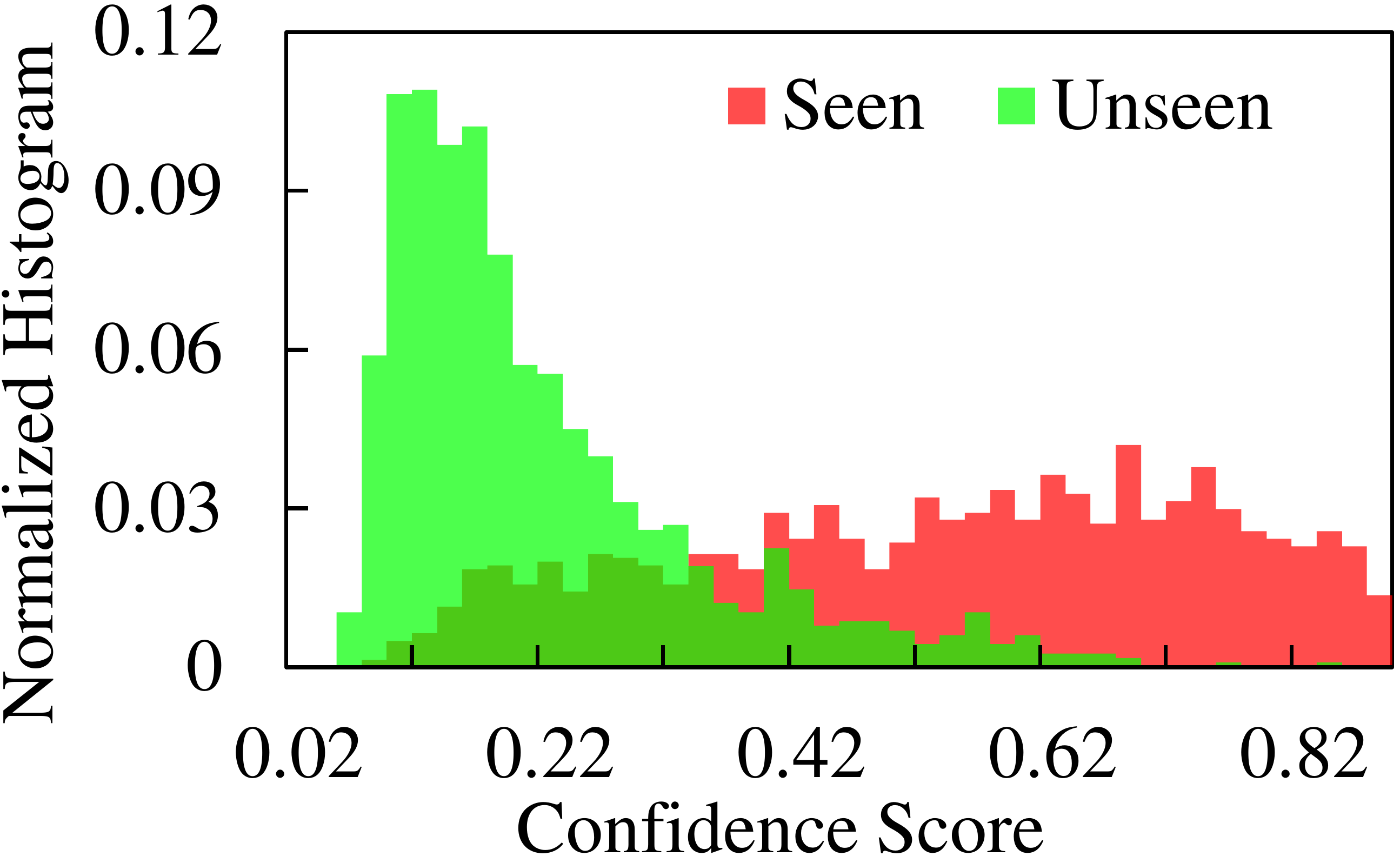}
  \end{minipage}} %

 \caption{Effectiveness of TC on domain segmentation.The left is the ROC curve. The upper right is the distribution of confidence scores without TC, and the bottom right is that with TC.}
\label{fig:3}
\end{figure}

The effect of TC on DS was also evaluated. For this purpose, we treated the unseen examples as the positive class, and the receiver operating characteristic curve (ROC) was taken as the evaluation protocol. In Fig. \ref{fig:3}, the left plot presents the ROC curve of the softmax classifier with and without TC for unseen detection. As shown in this figure, the classifier with TC can achieves higher true positive rate when the false positive rate is low. This is beneficial for domain segmentation in GZSL. Furthermore, the TC can reduce the sensitivity to the threshold for the detection of the unseen examples. The right of Fig.3 shows the distribution of confidence scores produced by the softmax classifier trained with (bottom) and without (upper) TC. It is evident that more than 64\% confidence scores of the seen classes are distributed between 0.98 and 1. This interval also contains the most confidence scores of unseen classes. As such, the model is sensitive to the threshold for distinguishing between the seen and unseen classes. This issue is effectively mitigated when the softmax classifier was trained with TC, as shown in the bottom right of Fig. \ref{fig:3}.

\subsection{Rethinking synthesized unseen data for GZSL}
Compared with these methods using generative models, the competitive result achieved by the current method warrants the consideration of the effectiveness of the augmented unseen samples on GZSL. To this end, we selected f-CLSWGAN \cite{xian2018feature} and dataset FLO as an example to analyze the effectiveness of the augmented data on GZSL. To be more general, instead of using the prototype-based embedding function, we used the simple embedding object function as follows:
\begin{equation}
    \mathop {\min }\limits_{_\psi } \sum\limits_{i = 1}^{N_s} {{{\left\| {\psi \left( {y_i^{(s)}} \right) - x_i^{\left( s \right)}} \right\|}^2}} 
\label{eq:embedding2}
\end{equation}
Table 3 exhibits GZSL performance of synthesized unseen data and CS. In this table, the baseline is represented by Eq. \ref{eq:embedding2}. Baseline + f-CLSWGAN \cite{xian2018feature} indicates that the unseen augmented data are included in the training set. As shown, compared with the baseline, the use of the augmented data significantly improves the performance of GZSL. Surprisingly, the simple implementation of CS achieves much better results compared with the use of the augmented data of f-CLSWGAN. 

\begin{table}[h]
\centering
\caption{GZSL performance on FLO with augmented unseen samples and CS. The results are measured in average per-class top-1 accuracy (\%).}
\label{tab:my-table3}
\setlength{\tabcolsep}{2.8mm}
\begin{tabular}{l|lll}
\toprule
& \multicolumn{3}{c}{FLO} \\ \hline
Method                        & ts     & tr     & H     \\ \hline
Baseline                      & 26.3   & 71.2   & 38.4  \\
Baseline + f-CLSWGAN   & 37.4   & 58.1   & 45.5  \\
Baseline + CS                 & 56.3   & 68.6   & 61.8  \\ \bottomrule
\end{tabular}
\label{tab:rethinking}
\end{table}

\begin{figure}
\centering
\subfigure[]{
   \begin{minipage}{0.3\columnwidth}
   \includegraphics[width=0.99\columnwidth]{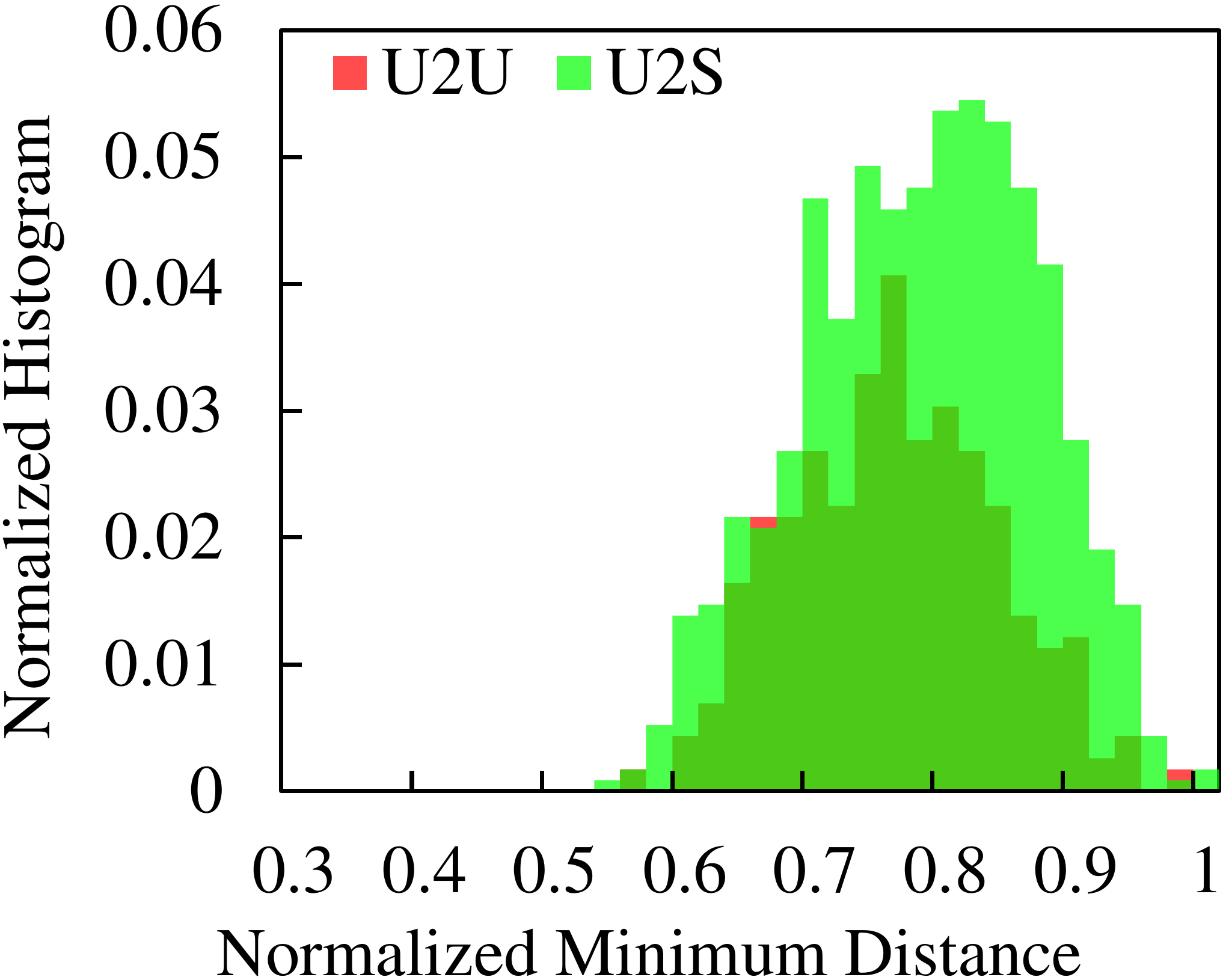}\\
   \includegraphics[width=0.99\columnwidth]{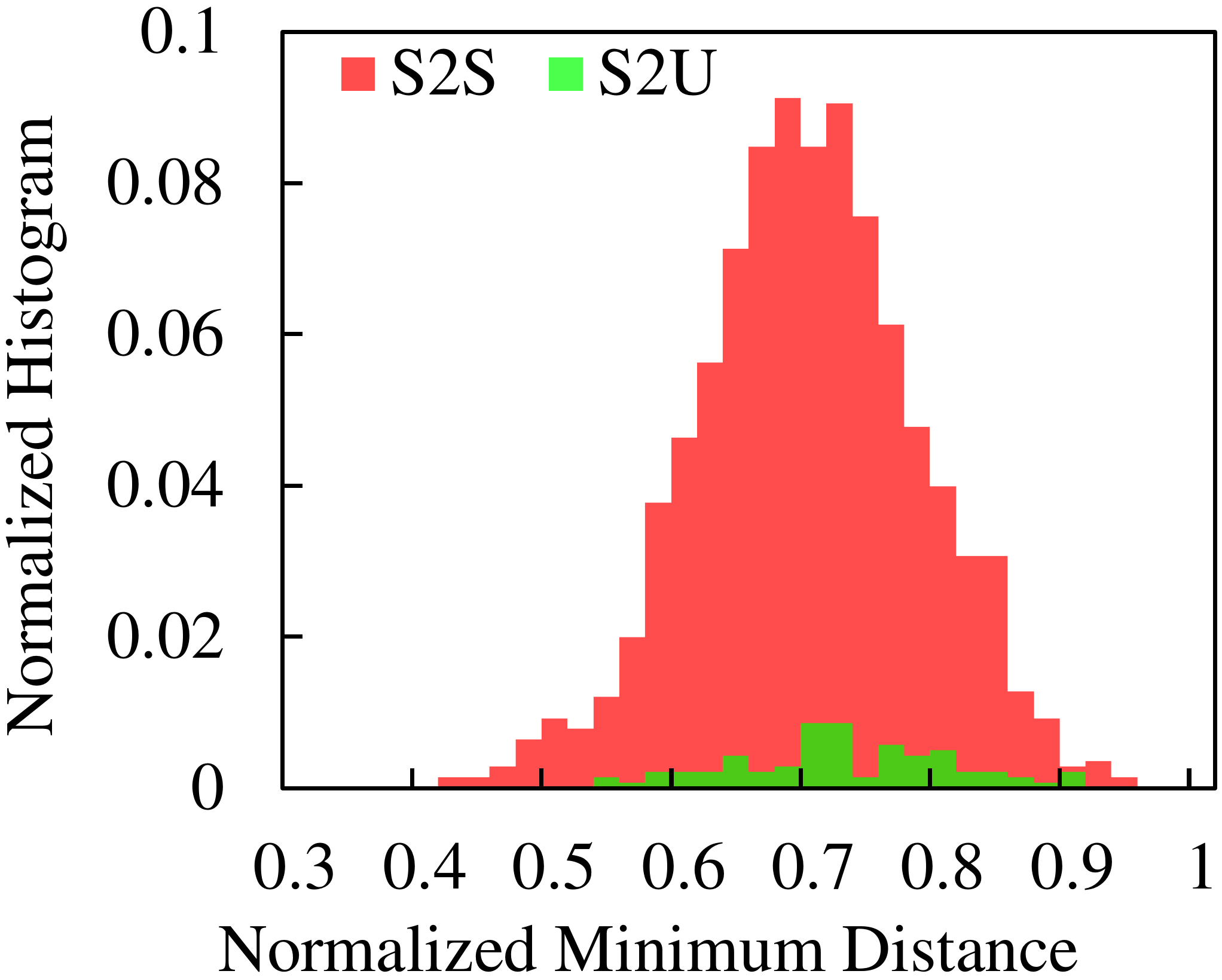}
  \end{minipage}
  \label{fig:4a}
  }
\subfigure[]{
   \begin{minipage}{0.3\columnwidth}
   \includegraphics[width=0.99\columnwidth]{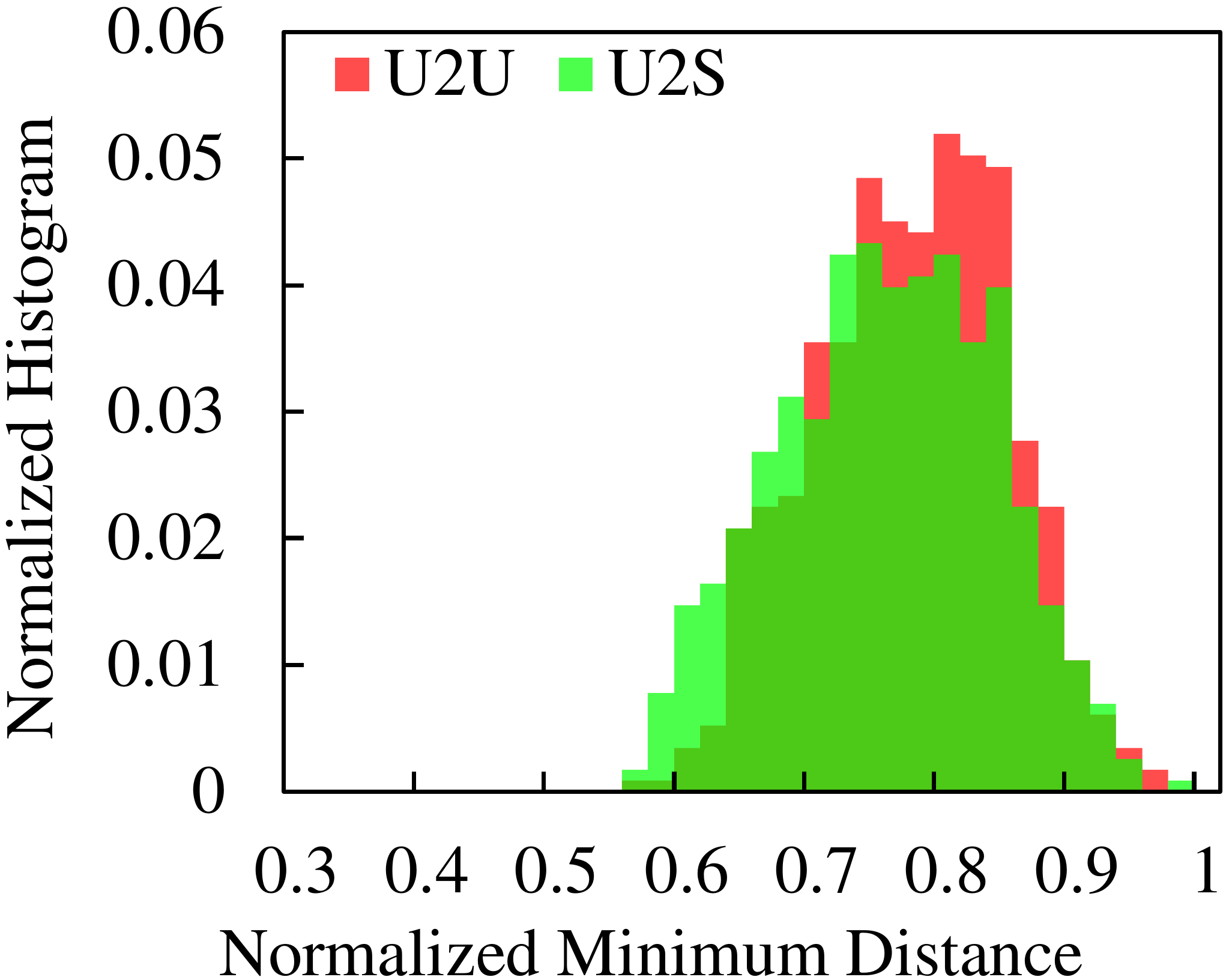}\\
   \includegraphics[width=0.99\columnwidth]{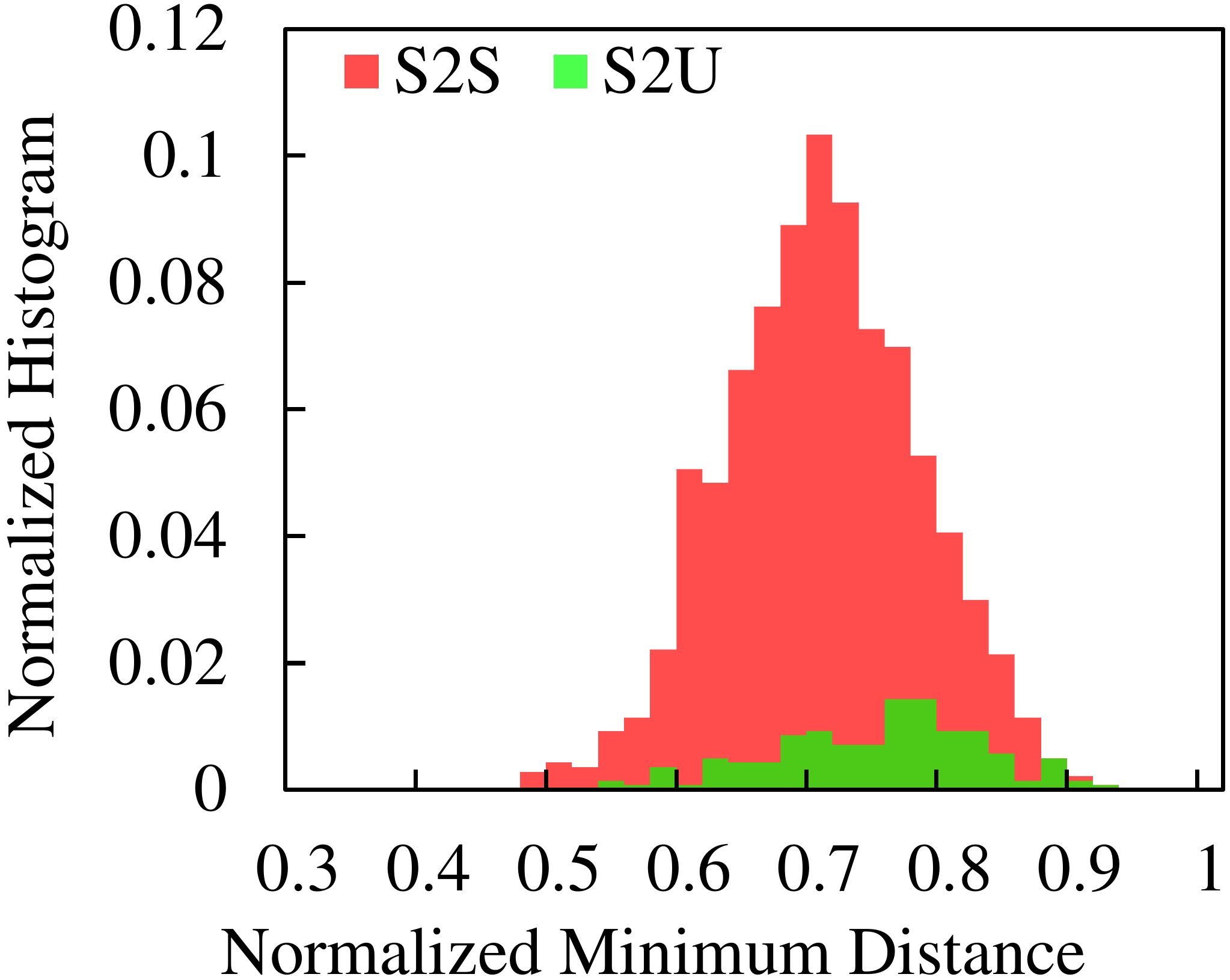}
  \end{minipage}
  }
 \subfigure[]{
   \begin{minipage}{0.3\columnwidth}
   \includegraphics[width=0.99\columnwidth]{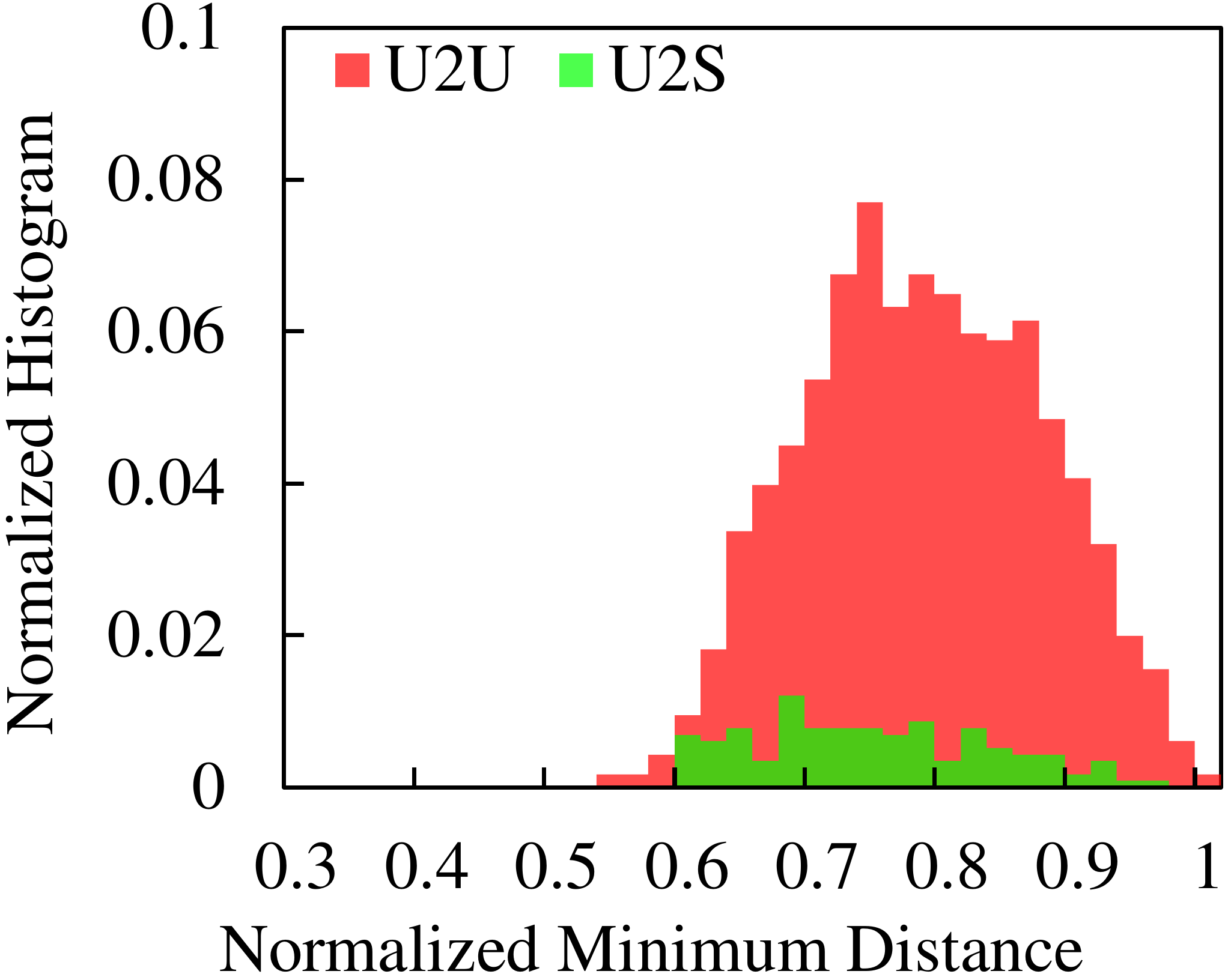}\\
   \includegraphics[width=0.99\columnwidth]{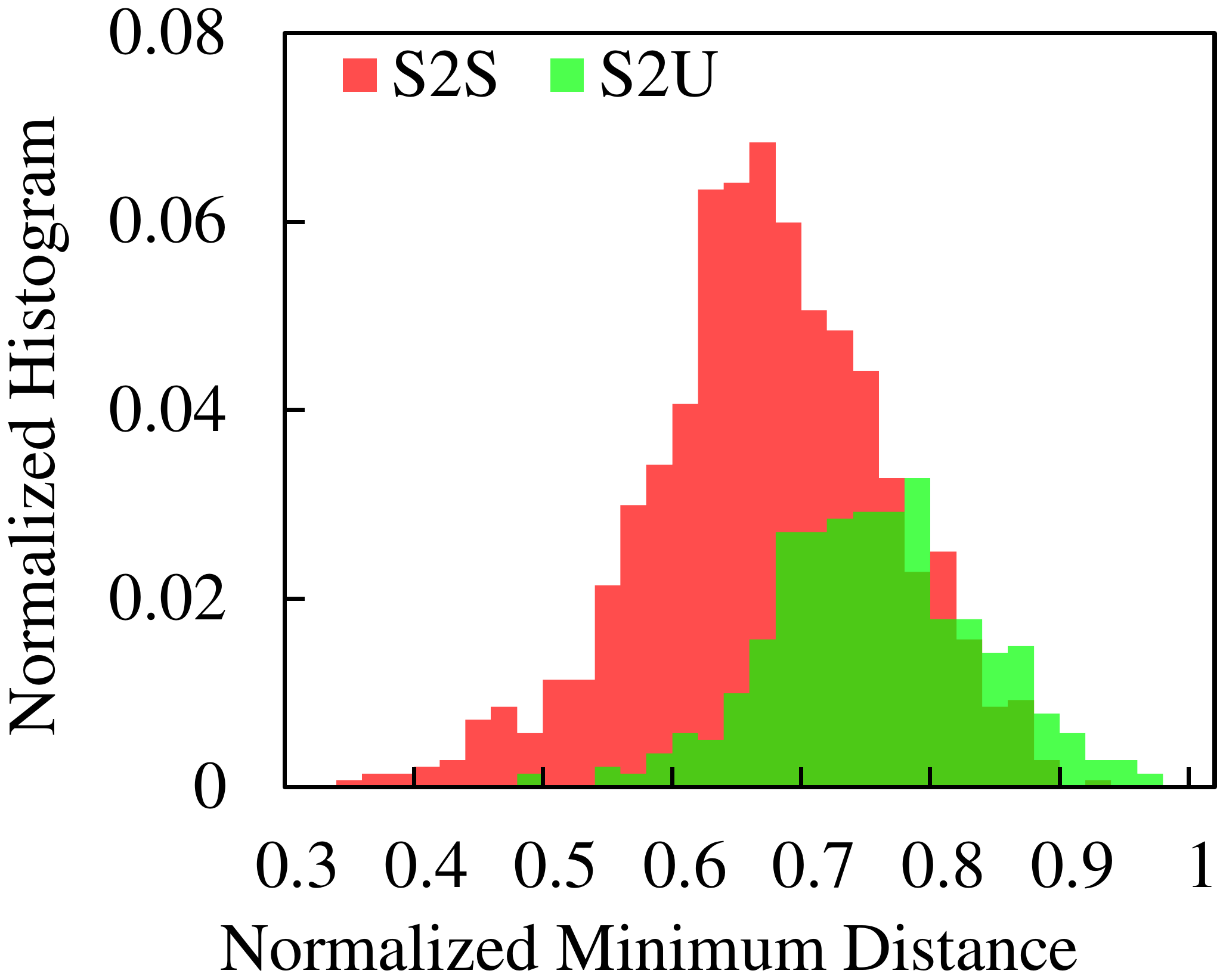}
  \end{minipage}
  }

 \caption{Distance distribution between visual features and predicted embedded semantic representations in the visual space. (a) Baseline. (b) Baseline + f-CLSWGAN. (c) Baseline + CS. In each group, the upper plot represents the distribution of unseen data, and the bottom plot represents the seen data. U and S represent unseen and seen classes, respectively. U2S refers to samples from unseen classes that are predicted as seen classes.}
\label{fig:4}
\end{figure}

The distributions of the predicted results are presented in Fig. \ref{fig:4}. In the baseline, most unseen instances are predicted as seen classes due to the domain shift problem. For the seen classes, only a few instances from the seen classes are incorrectly predicted as unseen classes. This situation is changed when the augmented unseen data are used during the training stage. Specifically, fewer unseen classes are predicted as seen classes, while more seen classes are predicted as unseen class. This change is further enhanced when CS is applied on the baseline, resulting in better performance on GZSL, as shown in Table. \ref{tab:rethinking}. This means that the augmented data may have a similar effect to CS on GZSL.

In the convolutional image recognition problem, augmented instances given by generative models can improve the abundance  of each class. However, in GZSL, the purpose of the generative models is to obtain unseen classes that never appeared in any previous training stage. The problem is that the generator in the generative model, to some extent, is also an embedding function trained with seen classes. Thus, the synthesized unseen features cause the classifier to pay less attention to seen classes instead of providing more latent information for unseen classes, compared with simple embedding-based classifiers. This contribution is most similar to that of CS, and the latter is simpler without accessing any unseen information.

\section{Conclusion}
In this work, we proposed to solve the domain shift problem in GZSL using domain segmentation. By combining the threshold of the softmax output and probabilistic distribution analysis, we successfully implemented a domain segmentation method to separate the training instances into seen, unseen, and uncertain domains. Among them, the uncertain domain is further adjusted by introducing CS. The proposed method yields outstanding performance on GZSL tasks and achieves new state-of-the-art results on five benchmark datasets. Moreover, we presented a comparison of the effects given by CS and the generative model on GZSL. The results indicate that the generative model may not be an ideal strategy for GZSL. On the one hand, generative models need access to the unseen semantic representations to generate unseen instances. On the other hand, the generator can only be trained with seen classes, which results in that the synthesized features fail to provide more abundant information, but it attempts to cause the classifier to pay less attention to the seen classes, which similar to the simple CS strategy. Thus, a better domain segmentation method or domain adjustment method should be more feasible for GZSL.

\bibliographystyle{aaai}
\bibliography{reference.bib}
\end{document}